\documentclass[letterpaper]{article} 
\usepackage{aaai2027}  
\nocopyright
\usepackage[hyphens]{url}  
\usepackage{graphicx} 
\usepackage{natbib}  
\usepackage{caption} 
\usepackage{xcolor}

\usepackage{amssymb}
\usepackage{subcaption}
\usepackage{amsmath}
\usepackage{algorithm}
\usepackage{algorithmic}

\usepackage{xspace}

\usepackage{newfloat}
\usepackage{listings}
\DeclareCaptionStyle{ruled}{labelfont=normalfont,labelsep=colon,strut=off} 
\floatstyle{ruled}
\newfloat{listing}{tb}{lst}{}
\floatname{listing}{Listing}

\usepackage{booktabs}

\title{PETA:Parameter-Efficient Test-Time Adaptation for Virtual Screening}
\author{
    Jia-Qi Lin\textsuperscript{\rm 1},
    Yinghua Yao\textsuperscript{\rm 1},
    Chang-Dong Wang\textsuperscript{\rm 2},
    Yew-Soon Ong\textsuperscript{\rm 1},
    Yuangang Pan\textsuperscript{\rm 1}\corresponding
}

\affiliations{
    \textsuperscript{\rm 1}Centre for Frontier AI Research (CFAR),
    Agency for Science, Technology and Research (A*STAR),
    Singapore 138632\\
    \textsuperscript{\rm 2}School of Computer Science and Engineering,
    Sun Yat-sen University, Guangzhou, China\\
    Lin\_Jiaqi@a-star.edu.sg,
    Yao\_Yinghua@a-star.edu.sg,\\
    changdongwang@hotmail.com,
    Ong\_Yew\_Soon@a-star.edu.sg,
    Pan\_Yuangang@a-star.edu.sg
}

\begin{document}

\maketitle

\begin{abstract}
Accurately ranking active ligands for a target protein pocket from massive chemical libraries remains a central challenge in virtual screening. DrugCLIP and its recent extensions substantially accelerate this process by encoding protein pockets and molecules into a shared embedding space. Despite this progress, further performance improvements typically require retraining the entire model, incurring substantial computational overhead and making target-specific customization inefficient.
In this work, we formulate the specialization of pretrained virtual screening models to individual pockets as a test-time adaptation problem and propose PETA, a parameter-efficient framework that directly adapts {pretrained model} at test time. 
Given a target pocket, PETA constructs pocket-specific negatives through molecular diffusion and chemical validity filtering, and further moves them toward the reference ligand retrieved from structural databases via embedding-space mixup to create more challenging ranking tasks. A ranking objective then places greater emphasis on suppressing high-scoring invalid candidates that could contaminate the top-ranked screening results, providing structured supervision for lightweight adaptation.
Experiments across diverse benchmarks demonstrate that this lightweight, pocket-specific adaptation outperforms both pretrained and fully retrained baselines while updating only the LayerNorm parameters, which account for approximately $0.03\%$ of the full model.
\end{abstract}

\section{Introduction}

Advances in chemical synthesis have expanded accessible chemical libraries to billions of diverse compounds~\cite{lyu2019ultra}. Although these libraries offer enormous opportunities for discovering new drug candidates, their scale makes exhaustive experimental screening impractical. Virtual screening~\cite{shoichet2004virtual} addresses this challenge by computationally prioritizing promising compounds for experimental validation and has therefore become an essential tool in early-stage drug discovery.

Existing virtual screening approaches can be broadly divided into docking-based and learning-based methods. Docking-based methods~\cite{alhossary2015fast,doi:10.1126/science.ads9530,DBLP:journals/jcc/TrottO10} search for plausible binding poses between a molecule and a protein pocket and use scoring functions to estimate their binding strength. The repeated conformational sampling and scoring required for each protein--molecule pair make these methods computationally expensive at scale. In contrast, learning-based approaches replace iterative docking procedures with efficient neural inference, directly predicting binding poses, binding affinities, or protein--ligand compatibility from data. 
Early learning-based methods~\cite{nguyen2021graphdta,singh2023contrastive} formulated virtual screening as binding-affinity regression or protein--ligand interaction classification, but their reliance on carefully labeled affinity data and reliable negative samples often limited generalization and performance. More recently, DrugCLIP~\cite{doi:10.1126/science.ads9530} reformulates virtual screening as a retrieval problem by independently encoding protein pockets and molecules into a shared embedding space, which replaces pairwise model inference with efficient vector similarity search, substantially accelerating large-scale virtual screening.

\begin{figure}[t]
    \centering
    \includegraphics[width=0.9\linewidth]{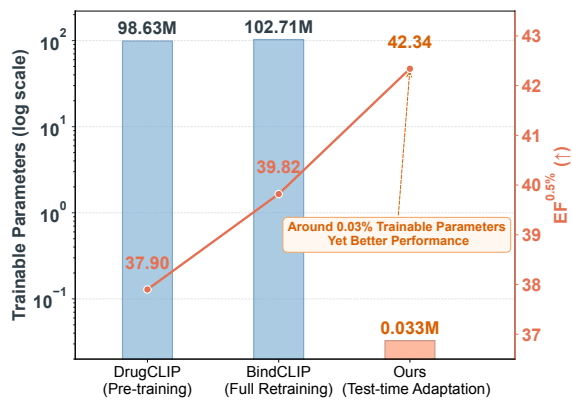}
  \caption{{PETA enables effective and parameter-efficient adaptation for virtual screening.}} 
    \label{fig:param}
\end{figure}


Following the retrieval paradigm established by DrugCLIP, recent studies have explored enhanced supervision and training strategies to {improve virtual screening performance on downstream unseen instances}~\cite{DBLP:journals/corr/abs-2602-15236,DBLP:conf/aaai/HanHL25,DBLP:conf/icml/ShenYMDHW25}. Despite their effectiveness, these approaches generally require rerunning the entire training pipeline, incurring substantial computational and data-processing costs. 
Which raises a key question: can an existing pretrained virtual screening model be adapted to a new target without full-model retraining?

In this work, we investigate whether a pretrained virtual screening model can be effectively and efficiently adapted to an unseen target pocket at test time. The key challenge is that how to obtain reliable adaptation signals for the unseen target pocket. To address this, we propose PETA, a parameter-efficient test-time adaptation framework that updates only LayerNorm parameters of pretrained model while constructing pocket-specific supervision on the fly. Specifically, PETA retrieves a structurally matched reference ligand for the target pocket and uses it as a positive anchor for adaptation. It further generates pocket-conditioned candidates, mines chemically invalid easy and hard negatives, and applies reference-guided embedding interpolation to create more challenging local comparisons. A cost-sensitive ranking objective prioritizes the correction of high-scoring invalid candidates, enabling pocket-specific adaptation by updating only the LayerNorm parameters. PETA updates only approximately $0.03\%$ of the full model parameters. Despite this lightweight update, it achieves an $\text{EF}^{0.5\%}$ of 42.34, outperforming the frozen DrugCLIP (37.90) and fully retrained BindCLIP (39.82), as shown in Figure~\ref{fig:param}. These results highlight a favorable balance between screening performance and adaptation efficiency.

Our main contributions are summarized as follows:
\begin{itemize}
    \item We formulate \emph{Test-Time Adaptation for Virtual Screening} (TTA-VS), which efficiently specializes a pretrained virtual screening model to each unseen protein pocket at inference time without binding annotations for the screening library or full-model retraining.
    
    \item We propose PETA, a parameter-efficient framework that constructs target-conditioned supervision from information available at test time while updating only a small fraction of the model parameters.
    
    \item Experiments across diverse benchmarks demonstrate the effectiveness and efficiency of PETA compared with both pretrained and full-retraining baselines.
\end{itemize}




\section{Related Works}

\paragraph{Drug Virtual Screening.}
Existing virtual screening methods can generally be divided into
docking-based and learning-based approaches. Docking-based methods, such as Glide-SP~\cite{friesner2004glide}, AutoDock
Vina~\cite{DBLP:journals/jcc/TrottO10}, and Surflex Dock~\cite{jain2003surflex}, search for plausible binding poses of small molecules within a target pocket and estimate their binding strengths using predefined scoring functions. Despite their broad applicability, these methods require repeated conformational sampling and pose evaluation for each protein--ligand pair, making large-scale screening computationally expensive.

Learning-based methods improve screening efficiency by replacing repeated conformational search and physics-based scoring with neural
inference~\cite{doi:10.1126/science.ads9530}. They learn protein--ligand interaction patterns from experimentally measured affinities or
structure-derived supervision and subsequently assign binding scores to compounds in a screening library. However, their performance remains constrained by the quality and coverage of available training data, as experimentally resolved complexes, reliable affinity measurements, and confirmed non-binding pairs are costly to obtain.

DrugCLIP~\cite{doi:10.1126/science.ads9530} mitigates some of these
limitations by learning a shared embedding space for protein pockets and ligands through contrastive learning. Virtual screening can then be performed through efficient embedding similarity search, reducing the reliance on affinity annotations and repeated pairwise model inference. Following this retrieval paradigm, AANet~\cite{DBLP:conf/nips/ZhuWGJTZML25} introduces tri-modal alignment and multi-cavity aggregation to support screening with apo and predicted protein structures. DrugHash~\cite{DBLP:conf/aaai/HanHL25} learns binary representations to reduce the memory and retrieval costs of ultra-large-scale screening, while BindCLIP~\cite{DBLP:journals/corr/abs-2602-15236} incorporates binding-pose generation and large-scale hard-negative mining to learn more discriminative pocket--ligand representations.
 

These methods primarily improve general-purpose representation learning or retrieval efficiency during offline training. Once a new target pocket is encountered, however, their scoring functions remain fixed unless the model is retrained with modified objectives or additional data. In contrast, PETA enables post-training specialization of an pretrained model to each unseen target pocket. It constructs pocket-specific ranking supervision from a retrieved reference ligand and pocket-conditioned negative candidates at test time, while updating only a small subset of model parameters.

\paragraph{Test-Time Adaptation.}
Test-time adaptation improves a pretrained model during inference without requiring labeled target-domain data and has been widely studied in image classification, semantic segmentation, and multimodal
learning~\cite{Wang2021,DBLP:conf/icml/NiuW0CZZT22,
DBLP:conf/cvpr/ChenPYLX24,DBLP:conf/nips/ShuNHYGAX22}. However, conventional test-time adaptation methods cannot be directly transferred to virtual screening. Rather than predicting labels for individual test samples, virtual screening requires ranking a large unlabeled ligand library for a specific target pocket, with particular emphasis on the highest-ranked compounds selected for experimental validation.

Drug-TTA~\cite{DBLP:conf/icml/ShenYMDHW25} is the most closely related work. It trains the entire model with multiple predefined auxiliary objectives and then performs additional instance-wise optimization during inference. This requires redesigning and rerunning the training pipeline before adaptation can be applied. PETA instead directly adapts an existing pretrained model by constructing pocket-specific ranking supervision on the fly and updating only its LayerNorm parameters.

\begin{figure*}[htpb]
    \centering
    \includegraphics[width=1\linewidth]{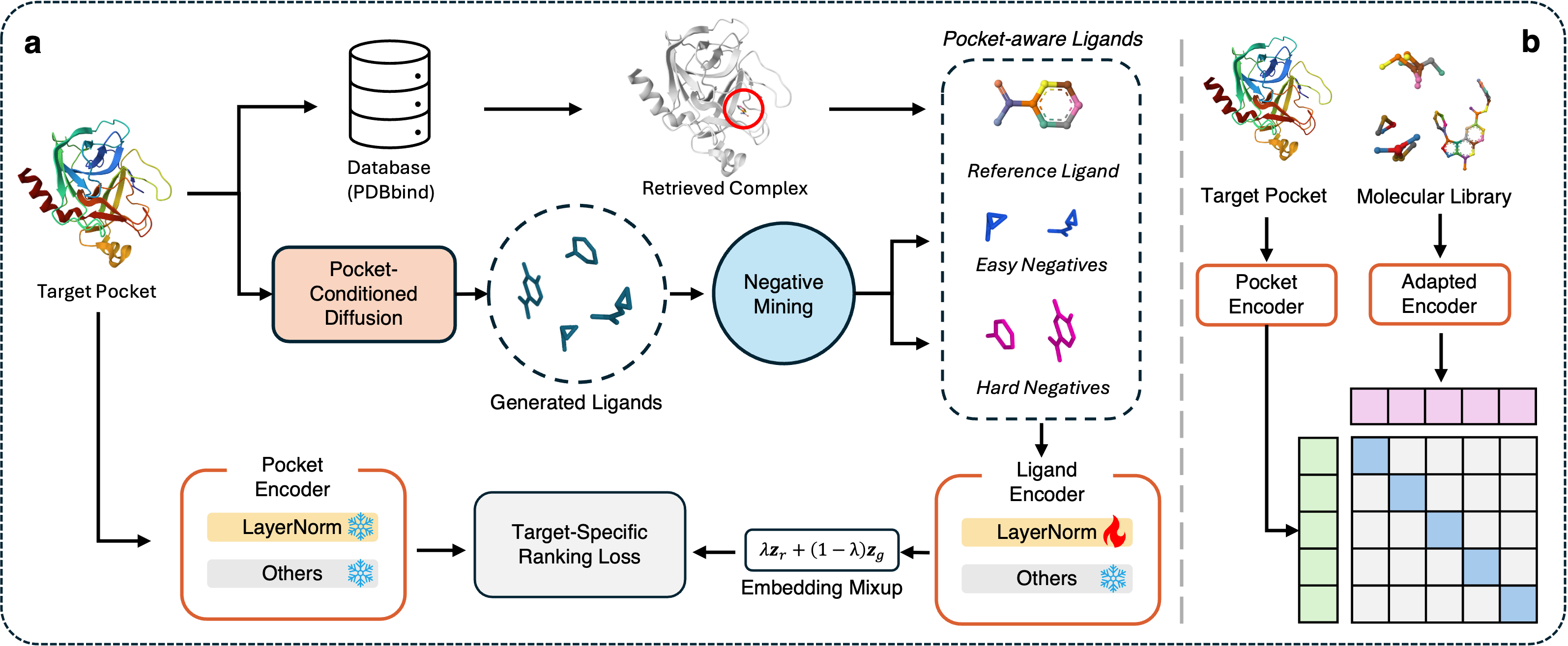}
    \caption{\label{fig:framework}Framework of PETA. {(a) Given a target pocket, PETA retrieves a reference ligand from a matched holo complex, constructs pocket-aware negatives, and adapts only the LayerNorm parameters of the ligand encoder. (b) The adapted model is then used for virtual screening over the test molecular library.}}
    
\end{figure*}

\section{Problem Formulation}
\label{sec:problem}

Given a target binding pocket $\mathbf{x}$ and a molecular screening library $\mathcal{M}=\{m_i\}_{i=1}^{N}$, virtual screening aims to rank the candidate ligands according to their predicted binding compatibility with $\mathbf{x}$ and identify the top-$K$ most promising compounds. Throughout this paper, \emph{ligand} refers to any molecule in the screening library, regardless of whether it is experimentally confirmed to bind to the target.

DrugCLIP performs virtual screening by encoding pockets and ligands into a
shared latent space. Let
\begin{equation}
\mathbf{z}_{\mathbf{x}} = \mathcal{E}_{\mathbf{x}}(\theta_{\mathbf{x}},\mathbf{x}),
\quad \mathbf{z}_{m} = \mathcal{E}_{m}(\theta_m,m),
\end{equation}
where $\mathcal{E}_{\mathbf{x}}$ and $\mathcal{E}_{m}$ denote the pocket and ligand encoders, respectively. The predicted binding score is defined as their cosine similarity:
\begin{equation}
s_{\theta}(\mathbf{x},m) =
\frac{\mathbf{z}_{\mathbf{x}}^{\top}\mathbf{z}_{m}}{
\|\mathbf{z}_{\mathbf{x}}\|\,\|\mathbf{z}_{m}\|},
\quad \theta=\{\theta_{\mathbf{x}},\theta_m\}.
\end{equation}
A higher score indicates stronger predicted pocket--ligand compatibility.

Although DrugCLIP is trained as a general-purpose screening model, its
performance may degrade on unseen or underrepresented target
pockets~\cite{DBLP:conf/icml/ShenYMDHW25}. Existing approaches typically
address this issue through additional objectives, datasets, and full-model
retraining, which is costly and does not directly specialize the model to each new target. We therefore study test-time specialization to an unseen pocket. The main challenge is to construct reliable pocket-specific supervision without affinity measurements or binding annotations for the screening library.

\section{Methodology}
\label{sec:method}
In this section, we propose \textbf{PETA}, a parameter-efficient test-time adaptation framework for virtual screening on unseen protein pockets. Rather than retraining the entire virtual screening model, PETA adapts the pretrained model independently for each target pocket by constructing pocket-specific supervision at inference time while updating only a small subset ($\approx 0.03\%$) of the model parameters. {Figure~\ref{fig:framework} illustrates the overall framework.}



\subsection{Test-Time Adaptation for Virtual Screening}
\label{sec:tta_vs}
Let $s_{\theta_0}(\mathbf{x},m)$ denote a pretrained virtual screening model with pretrained parameters $\theta_0$. At test time, given an unseen pocket $\mathbf{x}^{u}$ and a screening library $\mathcal{M}^{u}=\{m_i\}_{i=1}^{N}$ the model ranks the candidate ligands according to $ s_{\theta_0}(\mathbf{x}^{u},m_i)$. 

We formulate \emph{Test-Time Adaptation for Virtual Screening} (TTA-VS), which aims to specialize a pretrained model $s_{\theta_0}$ to an unseen pocket $\mathbf{x}^{u}$ without binding annotations for the screening library, additional affinity measurements, or full-model fine-tuning. 

\paragraph{Reference Ligand Construction.} Although $\mathbf{x}^{u}$ is unseen by the pretrained model, experimentally
resolved complexes involving the same protein target and binding site may be available in large structural databases\footnote{For all targets evaluated in this work, reference ligands were retrieved from PDBbind and excluded from the screening libraries.}. We therefore retrieve one such ligand as the reference ligand:
\begin{equation}\label{eq:reference_retrieval}
r_{\mathbf{x}^{u}} = \operatorname{Retrieve}
\left(\mathbf{x}^{u};\mathcal{D}_{\mathrm{struct}}\right),
\end{equation}
where $\mathcal{D}_{\mathrm{struct}}$ denotes a structural database.

A straightforward adaptation strategy is to maximize the similarity between the target pocket and the reference ligand:
\begin{equation} \label{eq:reference_only}
\mathcal{L}_{\mathrm{ref}} = - s_{\theta}
\left(\mathbf{x}^{u}, r_{\mathbf{x}^{u}} \right).
\end{equation}
However, this positive-only objective provides limited supervision for target-specific adaptation. It may increase the score of
$r_{\mathbf{x}^{u}}$, but fails to suppress false-positive candidates that already receive high scores. 
A natural solution is to augment the reference ligand with negative candidates (especially pocket-aware) $\mathcal{G}_{\mathbf{x}^{u}}$, thereby providing richer supervision for adaptation. The construction of informative negative candidates is described in detail in the following subsections.

To achieve parameter-efficient adaptation, the virtual screening model is then adapted by optimizing their relative ranking while updating only a small subset of parameters:
\begin{equation}  \label{eq:tta_objective}
\theta_a^{*} = \arg\min_{\theta_a} \mathcal{L} \left( \mathbf{x}^{u}, r_{\mathbf{x}^{u}} \cup\mathcal{G}_{\mathbf{x}^{u}}; \theta_a,\theta_b \right),  
\end{equation} 
where $\theta_a\subset\theta_0$ denotes the adaptable parameters and $\theta_b=\theta_0\setminus\theta_a$ remains frozen. The adapted model $\theta^{*}=(\theta_a^{*},\theta_b)$ is subsequently used to rank the screening library $\mathcal{M}^{u}$.

\subsection{Pocket-Conditioned Adaptation Set Construction}
\label{sec:pocket_negative}
\begin{figure}[t]
    \centering
    \includegraphics[width=0.9\linewidth]{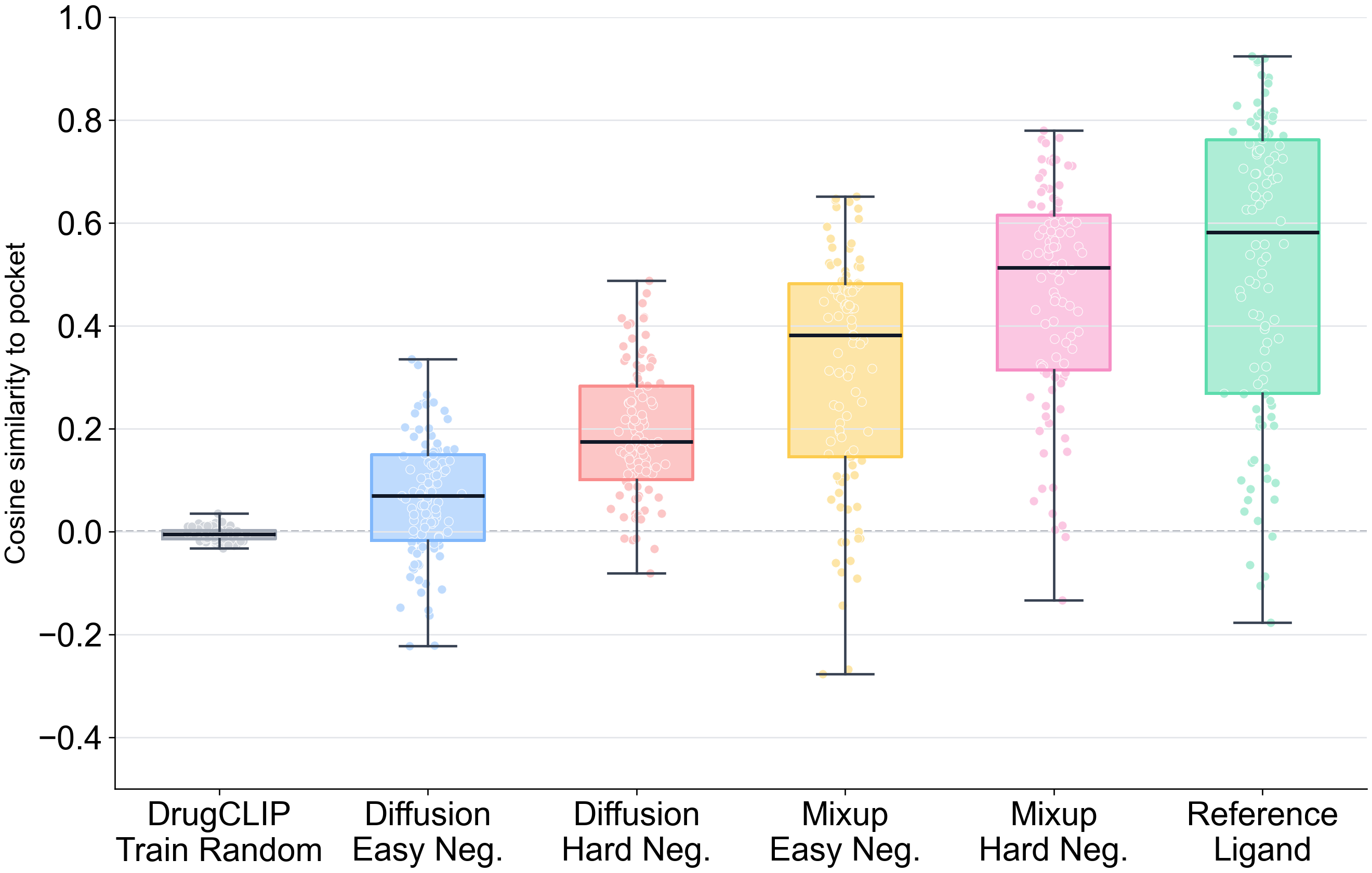}
    \caption{{Distributions of per-pocket average pocket--ligand cosine similarities on the DUD-E virtual screening benchmark.}}
    \label{fig:dude-box}
\end{figure}

One intuitive approach for constructing negative candidates is to randomly sample ligands from an existing molecular dataset. However, molecules sampled without considering the target pocket are generally unrelated to its local chemical environment and may already be easily distinguished by the pretrained model.
To examine this issue, we randomly sample 1,000 ligands from the DrugCLIP training set and compute their cosine similarities with target pockets in DUD-E using the pretrained model. As shown in Figure~\ref{fig:dude-box}, the resulting similarities are concentrated near zero. This indicates that most randomly sampled training ligands are already assigned low compatibility scores and therefore provide little useful gradient signal for adaptation. 

We instead construct candidates using a pocket-conditioned generative model
DiffSBDD $G_{\phi}$~\cite{DBLP:journals/ncs/SchneuingHDDJIDGBLWBC24}, parameterized by fixed parameters $\phi$.
Given an unseen target pocket $\mathbf{x}^{u}$, the generative model produces a candidate set:
\begin{equation} \label{eq:generated_set}
\mathcal{G}_{\mathbf{x}^{u}} = \{g_i\}_{i=1}^{B}, \quad g_i\sim G_{\phi}(g\mid\mathbf{x}^{u}).  
\end{equation}

\begin{figure}[t]
\centering
\includegraphics[width=0.75\linewidth]{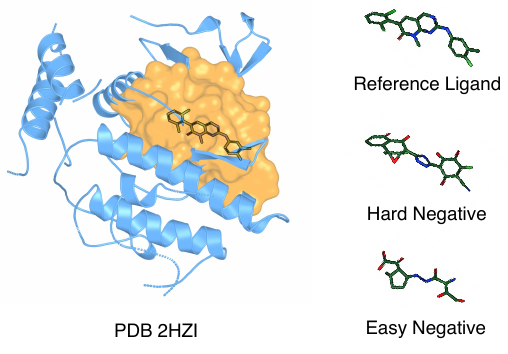}
\caption{\label{fig:ref_vs_neg} Reference ligand and pocket-conditioned candidates for ABL1. The crystal structure PDB 2HZI represents an experimentally resolved binding conformation of ABL1 in complex with the reference ligand PD180970. The generated candidates exhibit structural similarity to the reference ligand, providing pocket-relevant samples for adaptation.}
\end{figure}

Because generation is explicitly conditioned on the structural and physicochemical characteristics of the target binding site, the generated molecules explore a chemical space that is locally relevant to $\mathbf{x}^{u}$. Compared with randomly sampled training ligands, these candidates are more likely to exhibit structural patterns and molecular properties compatible with the target pocket.

As a result, pocket-conditioned candidates tend to be more difficult for the pretrained model to distinguish from plausible binders. They probe the local decision boundary of the pretrained embedding space and expose potentially overconfident or misleading predictions. Such candidates provide more informative gradients for pocket-specific adaptation than arbitrary molecules whose predicted similarities are already close to zero.

Figure~\ref{fig:ref_vs_neg} illustrates this property using ABL1. The generated candidates structurally resemble the known active reference ligand, suggesting that they occupy a chemically relevant neighborhood around the target pocket. However, structural relevance alone does not guarantee that a generated molecule is a valid binder. Some generated candidates may violate basic chemical constraints or exploit erroneous correlations learned by the pretrained model while still receiving high pocket--ligand similarity scores.

These observations motivate the subsequent negative mining procedure, which identifies candidate negatives from ${\mathcal{G}}_{\mathbf{x}^{u}}$ to form the final negative set $\mathcal{G}^{\mathrm{inv}}_{\mathbf{x}^{u}}$. Together with the reference ligand, the resulting negatives constitute a pocket-aware ligand set, enabling PETA to perform pocket-specific adaptation for the given unseen target pocket.



\subsection{Chemical Validity-Guided Negative Mining}
Pocket conditioning makes the generated ligands $\mathcal{G}_{\mathbf{x}^{u}}$ more relevant to the target pocket $\mathbf{x}^{u}$, but their binding activities remain unknown and some generated structures are chemically invalid. Generation alone therefore does not provide sufficiently reliable pseudo-labels.

We exploit chemical invalidity as an annotation-free source of negative supervision. A chemically invalid structure cannot represent a viable screening compound. More importantly, an invalid structure that nevertheless receives a high pocket--ligand similarity represents a high-risk false-positive prediction and provides an informative signal for adaptation.

For each generated ligand $g_i\in\mathcal{G}_{\mathbf{x}^{u}}$, we assess its chemical validity using the RDKit sanitization procedure~\cite{landrum2013rdkit}. Let $\mathrm{RD}(g_i)\in\{0,1\}$ denote the validity indicator, where $\mathrm{RD}(g_i)=1$
indicates that $g_i$ passes sanitization. We partition the generated ligand set into:
\begin{equation}\label{eq:val_inv}
\begin{aligned}
\mathcal{G}_{\mathbf{x}^{u}}^{\mathrm{val}}
&=\left\{g_i\in\mathcal{G}_{\mathbf{x}^{u}}\mid\mathrm{RD}(g_i)=1
\right\},\\
\mathcal{G}_{\mathbf{x}^{u}}^{\mathrm{inv}}
&=\left\{g_i\in\mathcal{G}_{\mathbf{x}^{u}}\mid\mathrm{RD}(g_i)=0
\right\}.
\end{aligned}
\end{equation}
Chemically valid candidates $\mathcal{G}_{\mathbf{x}^{u}}^{\mathrm{val}}$ are retained as plausible pocket-conditioned hypotheses, whereas invalid candidates $\mathcal{G}_{\mathbf{x}^{u}}^{\mathrm{inv}}$ provide reliable negative evidence.

Let $\mathbf{z}_{\mathbf{x}^{u}}$ and $\mathbf{z}_{g_i}$ denote the pocket and ligand representations produced by the pretrained model. The pocket--ligand similarity is:
\begin{equation}\label{eq:generated_similarity}
s_{\theta_0}(\mathbf{x}^{u},g_i) =
\frac{\mathbf{z}_{\mathbf{x}^{u}}^{\top}\mathbf{z}_{g_i}}{
\lVert\mathbf{z}_{\mathbf{x}^{u}}\rVert \lVert\mathbf{z}_{g_i}\rVert}.
\end{equation}

We then divide the invalid candidates $\mathcal{G}_{\mathbf{x}^{u}}^{\mathrm{inv}}$ into hard and easy negatives using the median similarity among all generated ligands $\mathcal{G}_{\mathbf{x}^{u}}$, namely,
\begin{equation}\label{eq:s_n}
\begin{aligned}
\mathcal{G}_{\mathbf{x}^{u}}^{\mathrm{inv\textrm{-}h}}
&=
\left\{g_i\in\mathcal{G}_{\mathbf{x}^{u}}^{\mathrm{inv}}
\mid s_{\theta_0}(\mathbf{x}^{u},g_i) \geq \gamma_{\mathbf{x}^{u}}
\right\},\\
\mathcal{G}_{\mathbf{x}^{u}}^{\mathrm{inv\textrm{-}e}}
&=
\left\{g_i\in\mathcal{G}_{\mathbf{x}^{u}}^{\mathrm{inv}}
\mid s_{\theta_0}(\mathbf{x}^{u},g_i) <\gamma_{\mathbf{x}^{u}}
\right\},
\end{aligned}
\end{equation}
where $\gamma_{\mathbf{x}^{u}} = Q_{0.5}
\left( \left\{
s_{\theta_0}(\mathbf{x}^{u},g_i) \mid g_i\in\mathcal{G}_{\mathbf{x}^{u}}
\right\} \right)$ is an adaptive threshold.

Easy negatives are chemically invalid structures that the pretrained model already assigns relatively low scores. In contrast, hard negatives are chemically invalid but receive scores in the upper half of the generated candidates. Although their geometric features may appear compatible with the pocket, they cannot represent viable ligands. Their high rankings therefore expose predictions that are particularly harmful to practical screening.

\subsection{Target-Specific Ranking Optimization}
Virtual screening is inherently top-heavy: only a small fraction of the highest-ranked compounds are typically selected for experimental validation. Consequently, a false positive near the top of the screening list is more consequential than a negative candidate that is already ranked low. Easy negatives provide complementary discrimination, whereas hard negatives represent high risk ranking errors that may displace viable compounds from the experimental shortlist.

\paragraph{Reference-Guided Latent Mixup.}
Let $\mathbf{z}_{r} = \mathcal{E}(r_{\mathbf{x}^{u}})$ and $\mathbf{z}_{g} = \mathcal{E}(g)$ denote the ligand representations of the reference $r_{\mathbf{x}^{u}}$ and a generated negative ligand $g \in \mathcal{G}_{\mathbf{x}^{u}}^{\mathrm{inv}}$,
respectively. Some generated negatives may remain far from the reference ligand in the pretrained representation space and thus yield trivial ranking constraints. To construct more challenging local comparisons, we interpolate each negative representation with the reference representation:
\begin{equation}\label{eq:mixup}
\widetilde{\mathbf{z}}_{g} = \lambda\mathbf{z}_{r} +
(1-\lambda)\mathbf{z}_{g},
\quad g\in \mathcal{G}_{\mathbf{x}^{u}}^{\mathrm{inv}},
\end{equation}
where $\lambda$ controls the interpolation strength and is set to $1/2$ in our experiments. 

The mixed representations lie closer to the reference ligand in the latent space, creating harder reference--negative comparisons. This prevents adaptation from relying only on trivially separable negatives and encourages the model to refine the ranking structure around the known binder.

\paragraph{Top-Heavy ListNet Objective.}
For each target pocket $\mathbf{x}^u$, we consider a candidate list consisting of one reference ligand $\mathbf{z}_r$ together with generated easy negatives and hard negatives:
\begin{equation}
\mathcal{C} = \{\mathbf{z}_r\} \cup \{\widetilde{\mathbf{z}}_{g}:g\in\mathcal{G}_{\mathbf{x}^{u}}^{\mathrm{inv\textrm{-}e}}\}
\cup\{\widetilde{\mathbf{z}}_{g}:g\in\mathcal{G}_{\mathbf{x}^{u}}^{\mathrm{inv\textrm{-}s}}\}.
\end{equation}

Motivated by the asymmetric ranking risk, we formulate a cost-sensitive
objective that places greater emphasis on correcting hard negatives. We
directly define the target probability distribution over the reference ligand,
easy negatives, and hard negatives as
\begin{equation}\label{eq:listnet_target}
P_y(c)=
\begin{cases}
\alpha, & c=\mathbf{z}_r,\\[1mm]
\dfrac{1-\alpha}
{|\mathcal{G}_{\mathbf{x}^{u}}^{\mathrm{inv\textrm{-}e}}|},
& c=\widetilde{\mathbf{z}}_g,\;
g\in\mathcal{G}_{\mathbf{x}^{u}}^{\mathrm{inv\textrm{-}e}},\\[2mm]
0,
& c=\widetilde{\mathbf{z}}_g,\;
g\in\mathcal{G}_{\mathbf{x}^{u}}^{\mathrm{inv\textrm{-}h}},
\end{cases}
\end{equation}
where $\alpha>0.5$. The reference ligand receives the largest probability mass, while the remaining mass is uniformly distributed among easy negatives. Hard negatives are assigned
zero target probability, reflecting their greater risk to the top-ranked
screening results.


For a candidate representation $c$, its pocket similarity is defined as $s_{\theta}(\mathbf{x}^{u},c)
= \frac{ \mathbf{z}_{\mathbf{x}^{u}}^{\top}c}{
\lVert\mathbf{z}_{\mathbf{x}^{u}}\rVert \lVert c\rVert}$. The predicted scores are converted into a probability distribution:
\begin{equation}\label{eq:listnet_prediction}
P_{\theta}(c\mid\mathbf{x}^{u})
=\frac{\exp\left(s_{\theta}(\mathbf{x}^{u},c)/\tau
\right)}{
\sum_{c'\in\mathcal{C}_{\mathbf{x}^{u}}}\exp\left(
s_{\theta}(\mathbf{x}^{u},c')/\tau\right)},
\end{equation}
where $\tau$ is the temperature parameter.

The target-specific ranking loss then defined by:
\begin{equation}\label{eq:listnet}
\mathcal{L}_{\mathrm{rank}}(\theta_a) = - \sum_{c\in\mathcal{C}_{\mathbf{x}^{u}}} P_y(c)
\log P_{\theta}(c\mid\mathbf{x}^{u}).
\end{equation}

The gradient with respect to a candidate score is:
\begin{equation}\label{eq:listnet_gradient}
\frac{\partial\mathcal{L}_{\mathrm{rank}}}
{\partial s_{\theta}(\mathbf{x}^{u},c)} = \frac{1}{\tau}
\left[ P_{\theta}(c\mid\mathbf{x}^{u})-P_y(c) \right].
\end{equation}
A hard negative has a relatively high predicted probability but zero target probability, resulting in a strong downward correction. Reference-guided embedding mixup further shifts these hard negatives toward the high-similarity region around the reference ligand, increasing their ambiguity and creating more challenging local ranking constraints. The resulting objective refines the local ranking structure and improves the reliability of the top-ranked screening candidates.

\subsection{Parameter-Efficient Pocket-Specific Adaptation}
\label{sec:parameter_efficient_adaptation}

For each target pocket, PETA updates only the LayerNorm of the ligand encoder, while the pocket encoder, the remaining ligand-encoder parameters, and the generator $G_{\phi}$ remain frozen. We optimize Eq.~\eqref{eq:listnet} for thirty steps and then use the adapted model to rank all compounds in $\mathcal{M}^{u}$:
\begin{equation}
\operatorname{rank} \left( \left\{s_{\theta^{*}}(\mathbf{x}^{u},m_i)
\right\}_{m_i\in\mathcal{M}^{u}}\right),
\end{equation}
where $\theta^*$ denotes the parameter of adapted model.
After screening one pocket, the adaptable parameters are reset to their pretrained initialization before processing the next pocket. This episodic procedure prevents information transfer across target pockets and ensures that each adaptation is independently conditioned on the corresponding pocket and reference ligand. The overall procedure is summarized in Algorithm~\ref{alg:ltta}.

\begin{algorithm}[t]\caption{PETA for virtual screening} 
\label{alg:ltta}
\textbf{Require:} Pocket encoder $\mathcal{E}_\mathbf{x}(\theta_\mathbf{x},\cdot)$, ligand encoder $\mathcal{E}_\mathbf{m}(\theta_m,\cdot)$, target pocket set $\mathcal X$.
\begin{algorithmic}[1]
\FOR{each target pocket $\mathbf{x}^u\in\mathcal P$}
\STATE Retrieve the reference ligand $r_{\mathbf{x}^{u}}$ by Eq.~\eqref{eq:reference_retrieval}
\STATE Encode the target pocket and reference ligand: $\mathbf z_{\mathbf{x}^u},\mathbf z_{r}$. 
\STATE Generate candidate ligands $\mathcal{G}_{\mathbf{x}^{u}}$ by Eq.~\eqref{eq:generated_set}.
\STATE Construct invalid ligands set $\mathcal{G}_{\mathbf{x}^{u}}^{\mathrm{inv}} = \mathcal{G}_{\mathbf{x}^{u}}^{\mathrm{inv\textrm{-}e}}\cup \mathcal{G}_{\mathbf{x}^{u}}^{\mathrm{inv\textrm{-}h}}$ 
 according to Eq.~\eqref{eq:val_inv} and Eq.~\eqref{eq:s_n}.
\FOR{each $g\in\mathcal{G}_{\mathbf{x}^{u}}^{\mathrm{inv}}$}
\STATE Construct the mixed embedding $\widetilde{\mathbf{z}}_{g}$ by Eq.~\eqref{eq:mixup}.
\ENDFOR
\STATE Update the LayerNorm of $\theta_m$ with $\mathcal L_{\mathrm{rank}}$ (Eq.~\eqref{eq:listnet}).
\STATE Screen pocket $\mathbf{x}^u$ using the adapted model.
\STATE Reset the LayerNorm of $\theta_m$ to its pretrained state.
\ENDFOR
\end{algorithmic}
\end{algorithm}

\section{Experiments}


\subsection{Experimental Settings}

\subsubsection{Datasets.}
To comprehensively evaluate the effectiveness of our method, we conduct experiments on two widely used virtual screening benchmarks, namely DUD-E~\cite{mysinger2012directory} and LIT-PCBA~\cite{tran2020lit}.
{Furthermore, following the Boltz-2 evaluation protocol, we assess fine-grained ligand ranking on the four-target FEP+ benchmark~\cite{ross2023maximal} comprising CDK2, TYK2, JNK1, and P38~\cite{hahn2022best}. This benchmark evaluates relative affinity ranking among structurally related ligands, requiring sensitivity to subtle chemical modifications.}

For reference ligands, they can be obtained from structural databases such as PDBbind~\cite{liu2017forging}, BioLiP2~\cite{zhang2024biolip2}, and PLINDER~\cite{durairaj24plinder}. In our experiments, all reference ligands are retrieved from PDBbind by matching each evaluation pocket to an experimentally determined holo complex of the same target and binding site. 

\paragraph{Baselines.}
On DUD-E, we compare with AutoDock Vina~\cite{DBLP:journals/jcc/TrottO10} and Glide-SP~\cite{friesner2004glide} as docking baselines, and RF-Score~\cite{DBLP:journals/bioinformatics/BallesterM10}, Pafnucy~\cite{stepniewska2017pafnucy}, OnionNet~\cite{zheng2019onionnet}, PLANET~\cite{zhang2023planet}, DrugCLIP~\cite{doi:10.1126/science.ads9530}, DrugHash~\cite{DBLP:conf/aaai/HanHL25}, and BindCLIP~\cite{DBLP:journals/corr/abs-2602-15236} as learning-based baselines. On LIT-PCBA, the baselines include Surflex~\cite{spitzer2012surflex} and Glide-SP for docking, and PLANET, GNINA~\cite{DBLP:journals/jcheminf/McNuttFAMMRSK21}, DeepDTA~\cite{ozturk2018deepdta}, BigBind~\cite{DBLP:journals/jcisd/BrocidiaconoFAPKT24}, DrugCLIP, DrugHash, and BindCLIP for learning-based screening. On FEP+, we compare with DrugCLIP and BindCLIP.

\paragraph{Evaluation Metrics.}

Virtual screening performance is evaluated using the area under the ROC curve (AUROC), Boltzmann-enhanced discrimination of ROC (BEDROC), and enrichment factor (EF).
For the FEP+ benchmark, we evaluate fine-grained ranking using pairwise accuracy and Kendall rank correlation $\mathcal{K}$. Pairwise accuracy measures the proportion of correctly ordered ligand pairs according to experimental $\Delta\Delta G$ relations, while $\mathcal{K}$ assesses agreement between the predicted and experimental global rankings.



\subsection{Evaluation on Benchmarks}

\paragraph{Virtual Screening.} Results on DUD-E and LIT-PCBA are reported in Tables~\ref{tab:dude} and~\ref{tab:lit-pcba}, respectively. PETA achieves the best BEDROC and EF results on both benchmarks. In particular, compared with the pretrained DrugCLIP, PETA improves $\text{EF}^{0.5\%}$ by 4.44 on DUD-E and 3.38 on LIT-PCBA. PETA also consistently outperforms the fully trained DrugHash and BindCLIP across all early-enrichment metrics, demonstrating that target-specific adaptation can provide stronger screening performance without retraining the entire model. On DUD-E, PETA further achieves the best AUROC, BEDROC, and EF values across all evaluated methods.
On LIT-PCBA, PETA achieves an AUROC of 57.56\%, which is lower than GNINA (60.93\%) and BindCLIP (59.15\%). Nevertheless, PETA obtains the best BEDROC of 8.84, surpassing BindCLIP (7.88) and DrugCLIP (6.41), and also achieves the highest EF at all evaluated cutoffs. This difference reflects the distinct focuses of the metrics: AUROC measures discrimination over the complete ranked library, whereas BEDROC and EF emphasize the early retrieval of active compounds. Since practical virtual screening typically selects only a small top-ranked fraction for experimental validation, early-enrichment metrics are particularly relevant. These results indicate that PETA more effectively concentrates active ligands near the top of the ranking.



\begin{figure}[htpb!]
    \centering
    \begin{subfigure}[t]{0.48\linewidth}
        \centering
        \includegraphics[width=\linewidth]{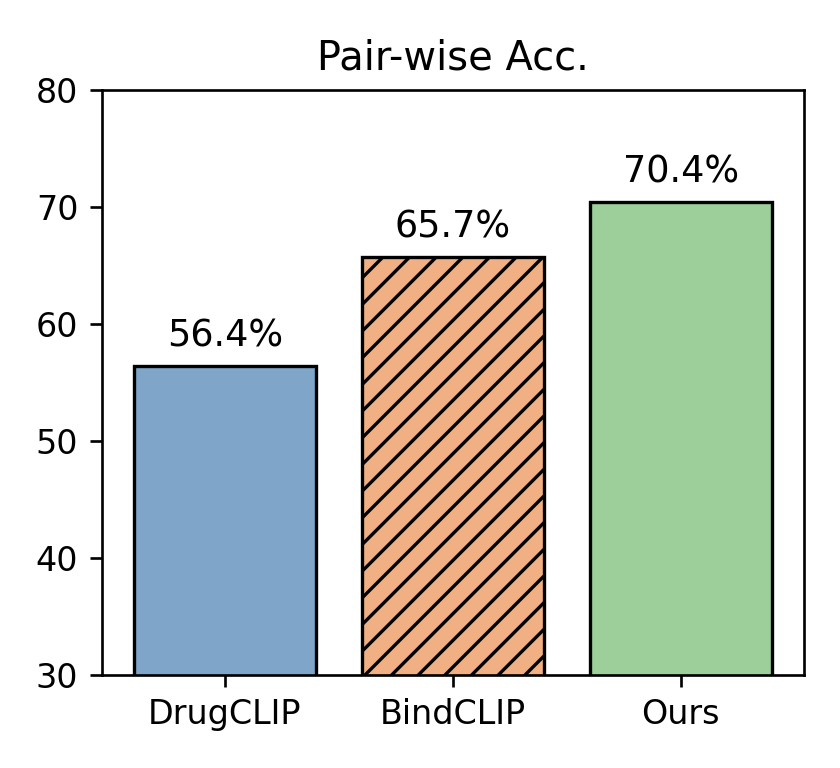}
    \end{subfigure}
    \begin{subfigure}[t]{0.48\linewidth}
        \centering
        \includegraphics[width=\linewidth]{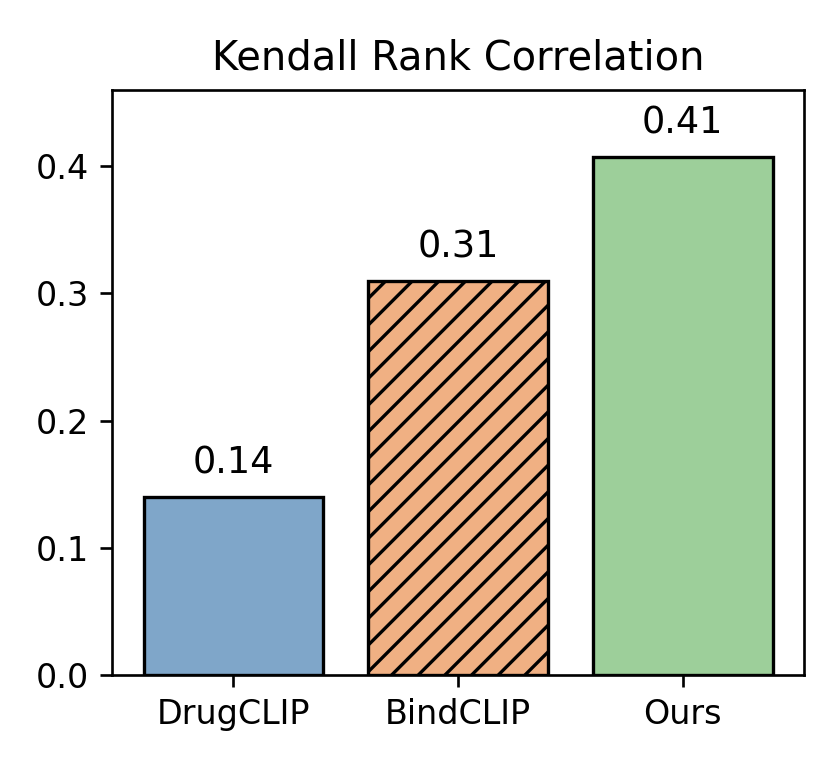}
    \end{subfigure}
    \caption{Fine-grained ligand-ranking performance on the four-target FEP+ benchmark. Higher values indicate better performance.}
    \label{fig:fep}
\end{figure}

\paragraph{Fine-Grained Ligand Ranking.}
Figure~\ref{fig:fep} reports results on the four-target FEP+ benchmark, comparing predicted ligand rankings with the relative affinity orderings derived from experimental $\Delta\Delta G$ relations. PETA performs best on both pairwise accuracy and $\mathcal{K}$, demonstrating stronger discrimination among closely related ligands.
PETA achieves 70.4\% pairwise accuracy, exceeding BindCLIP and DrugCLIP by 4.7 and 14.0 percentage points, respectively. It also obtains a Kendall rank correlation of 0.41, compared with 0.31 and 0.14. These gains show that target-specific adaptation improves both pairwise affinity comparisons and globally consistent ranking within congeneric ligand series.

\begin{table}[htpb!]
\centering
\resizebox{\linewidth}{!}{
\begin{tabular}{lccccc}
\toprule
Method & AUROC & BEDROC & EF$^{0.5\%}$ & EF$^{1\%}$ & EF$^{5\%}$ \\
\midrule
Vina & 71.60 & -- & 9.13 & 7.32 & 4.44 \\
Glide-SP & 76.70 & 40.70 & 19.39 & 16.18 & 7.23 \\
\midrule
RFscore & 65.21 & 12.41 & 4.90 & 4.52 & 2.98 \\
Pafnucy & 63.11 & 16.50 & 4.24 & 3.86 & 3.76 \\
OnionNet & 59.71 & 8.62 & 2.84 & 2.84 & 2.20 \\
PLANET & 71.60 & -- & 10.23 & 8.83 & 5.40 \\
DrugHash & 80.05 & 47.22 & 37.28 & 29.57 & 9.37 \\
DrugCLIP & 79.29 & 47.53 & 37.90 & 30.52 & 10.08 \\
BindCLIP & 80.14 & 49.73 & 39.82 & 32.16 & 10.44 \\
\midrule
Ours & \textbf{82.62} & \textbf{53.86} & \textbf{42.34} & \textbf{34.65} & \textbf{11.44} \\
\bottomrule
\end{tabular}}
\caption{Virtual screening results on DUD-E. The best results are highlighted in \textbf{bold}.}
\label{tab:dude}
\end{table}

\begin{table}[htpb!]
\centering
\resizebox{\linewidth}{!}{
\begin{tabular}{lccccc}
\toprule
Method & AUROC & BEDROC & EF$^{0.5\%}$ & EF$^{1\%}$ & EF$^{5\%}$ \\
\midrule
Surflex & 51.47 & -- & -- & 2.50 & -- \\
Glide-SP & 53.15 & 4.00 & 3.17 & 3.41 & 2.01 \\
\midrule
PLANET & 57.31 & -- & 4.64 & 3.87 & 2.43 \\
GNINA & \textbf{60.93} & 5.40 & -- & 4.63 & -- \\
DeepDTA & 56.27 & 2.53 & -- & 1.47 & -- \\
BigBind & 60.80 & -- & -- & 3.82 & -- \\
DrugHash & 54.34 & 6.49 & 7.86 & 5.35 & 2.32 \\
DrugCLIP & 55.45 & 6.41 & 8.24 & 5.21 & 2.14 \\
BindCLIP & 59.15 & 7.88 & 9.84 & 6.26 & 2.90 \\
\midrule
Ours & 57.56 & \textbf{8.84} & \textbf{11.62} & \textbf{7.29} & \textbf{3.22} \\

\bottomrule
\end{tabular}}
\caption{Virtual screening results on LIT-PCBA. The best results are highlighted in \textbf{bold}.}
\label{tab:lit-pcba}
\end{table}

\begin{table}[t]
\centering
\resizebox{\linewidth}{!}{
\begin{tabular}{cccc|ccccc}
\toprule
$r_{\mathbf{x}^u}$ & $\mathcal{G}_{\mathbf{x}^u}^{\mathrm{inv\textrm{-}e}}$ & $\mathcal{G}_{\mathbf{x}^u}^{\mathrm{inv\textrm{-}h}}$ & \textsc{Mixup}
& AUROC & BEDROC & EF$^{0.5\%}$ & EF$^{1\%}$ & EF$^{5\%}$ \\
\midrule
$\times$     & $\checkmark$ & $\checkmark$ & $\checkmark$
& 80.16 & 48.11 & 38.49 & 31.50 & 10.64 \\

$\checkmark$ & $\times$     & $\times$     & $\checkmark$
& {81.23} & 51.74 & 40.41 & {31.90} & 10.80 \\

$\checkmark$ & $\times$     & $\checkmark$ & $\checkmark$
& 81.48 & 51.59 & 40.05 & 32.52 & 10.89 \\

$\checkmark$ & $\checkmark$ & $\times$     & $\checkmark$
& 81.23 & 51.80 & 40.55 & 32.82 & 11.02 \\

$\checkmark$ & $\checkmark$ & $\checkmark$ & $\times$
& 81.32 & 51.29 & 41.12 & 33.18 & 11.35 \\
\midrule
$\checkmark$ & $\checkmark$ & $\checkmark$ & $\checkmark$
& \textbf{82.62} & \textbf{53.86} & \textbf{42.34} & \textbf{34.65} & \textbf{11.44} \\
\bottomrule
\end{tabular}}
\caption{Ablation studies on DUD-E. The best results are highlighted in \textbf{bold}.}
\label{tab:ablation-dude}
\end{table}

\begin{table}[t]
\centering
\resizebox{\linewidth}{!}{
\begin{tabular}{cccc|ccccc}
\toprule
$r_{\mathbf{x}^u}$ & $\mathcal{G}_{\mathbf{x}^u}^{\mathrm{inv\textrm{-}e}}$ & $\mathcal{G}_{\mathbf{x}^u}^{\mathrm{inv\textrm{-}h}}$ & \textsc{Mixup}
& AUROC & BEDROC & EF$^{0.5\%}$ & EF$^{1\%}$ & EF$^{5\%}$ \\
\midrule
$\times$     & $\checkmark$ & $\checkmark$ & $\checkmark$
& 55.60 & 7.15 & 8.07 & 5.32 & 2.29 \\

$\checkmark$ & $\times$     & $\times$     & $\checkmark$
& 57.66 & 7.51 & 9.08 & 6.62 & 2.86 \\

$\checkmark$ & $\times$     & $\checkmark$ & $\checkmark$
& 58.02 & 7.82 & 9.29 & 6.99 & {3.09} \\

$\checkmark$ & $\checkmark$ & $\times$     & $\checkmark$
& \textbf{59.44} & 7.69 & 9.17 & 6.53 & 3.06 \\

$\checkmark$ & $\checkmark$ & $\checkmark$ & $\times$
& 58.03 & 7.88 & 9.73 & {7.12} & 2.97 \\
\midrule
$\checkmark$ & $\checkmark$ & $\checkmark$ & $\checkmark$
& 57.56 & \textbf{8.84} & \textbf{11.62} & \textbf{7.29} & \textbf{3.22} \\
\bottomrule
\end{tabular}}
\caption{Ablation studies on LIT-PCBA. The best results are highlighted in \textbf{bold}.}
\label{tab:ablation-lit-pcba}
\end{table}

\subsection{Ablation Studies}

\paragraph{Component Ablation.}
We examine four components of PETA: the reference ligand $r_{\mathbf{x}^u}$, the easy negative set $\mathcal{G}_{\mathbf{x}^u}^{\mathrm{inv\textrm{-}e}}$, the hard negative set $\mathcal{G}_{\mathbf{x}^u}^{\mathrm{inv\textrm{-}h}}$, and embedding space mixup. To ablate the reference ligand, we replace it with a randomly selected valid candidate from $\mathcal{G}_{\mathbf{x}^u}^{\mathrm{val}}$, thereby removing the known-active anchor while preserving the adaptation procedure. To assess hierarchical negative mining, we remove the easy negative and hard negative sets individually and jointly. When neither set is used, we instead sample 50 random ligands from the DrugCLIP training set as negatives. To ablate mixup, we apply the ranking objective directly to the original ligand embeddings. All other experimental settings remain unchanged. Results on DUD-E and LIT-PCBA are reported in Tables~\ref{tab:ablation-dude} and~\ref{tab:ablation-lit-pcba}, respectively.
From these tables, we observe that no partial configuration consistently performs best across both datasets. The complete configuration achieves the highest BEDROC on both DUD-E and LIT-PCBA while remaining competitive on the other metrics, demonstrating a balanced overall performance.

\begin{figure}[ht]
\begin{subfigure}[htpb!]{0.48\linewidth}
\centering
\includegraphics[width=\linewidth]{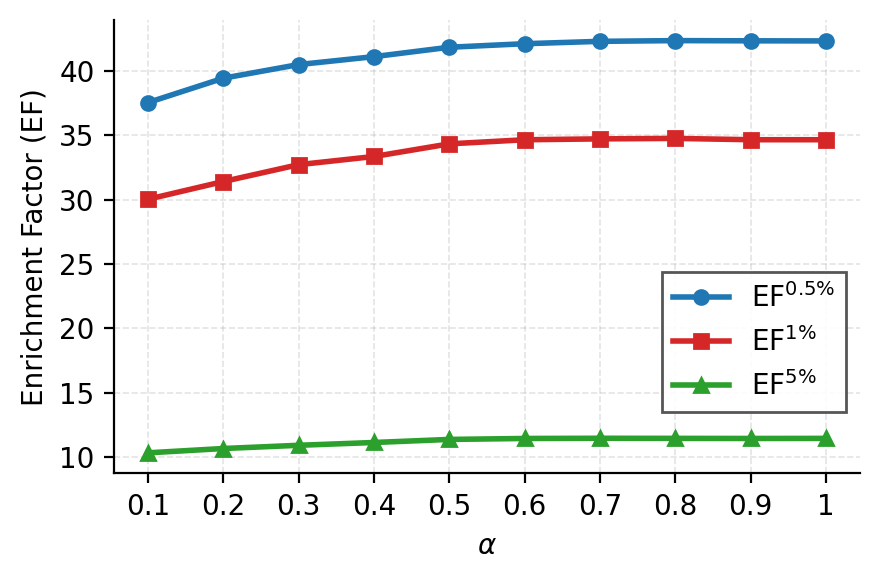}
\caption{DUD-E.}
\label{fig:dude_alpha}
\end{subfigure}
\hfill
\begin{subfigure}[htpb!]{0.48\linewidth}
\centering
\includegraphics[width=\linewidth]{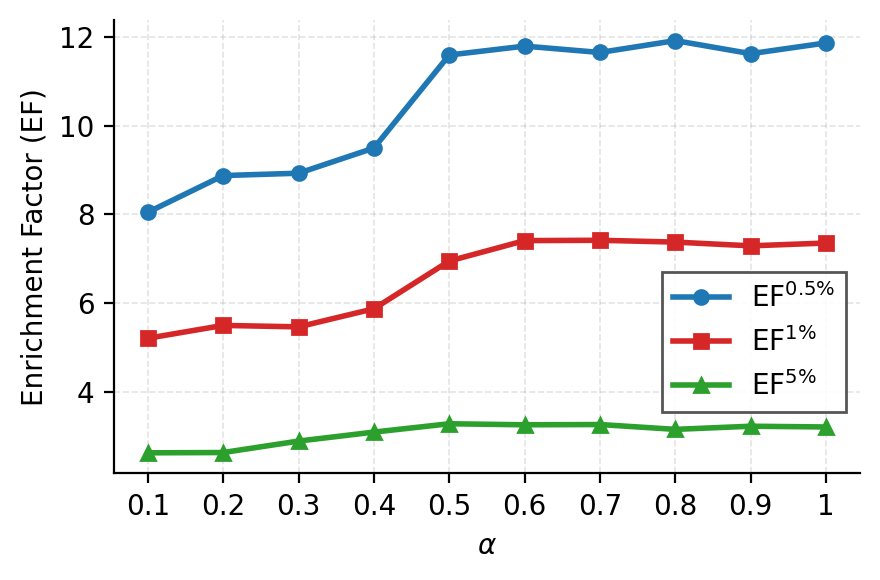}
\caption{LIT-PCBA.}
\label{fig:pcba_alpha}
\end{subfigure}

\caption{Ablation of $\alpha$ on virtual screening performance.}
\label{fig:weight-allocation}
\end{figure}

\paragraph{Effect of Target-Specific Ranking Weight.}
We examine the effect of $\alpha$ by varying it from {0.1 to 1} in increments of 0.1 in Eq.~\eqref{eq:listnet}. Results are shown in Figure~\ref{fig:weight-allocation}.
From these figures, we observe that both DUD-E and LIT-PCBA exhibit a similar trend. The known active reference ligand serves as a reliable positive anchor, defining the desired target-specific optimization direction in the embedding space. When $\alpha$ is small, the negative candidates dominate the ranking objective, potentially steering the parameter updates away from this desired direction and degrading screening performance. As $\alpha$ increases, the reference ligand exerts greater influence on adaptation, bringing the parameter updates into better alignment with the desired ranking direction. Performance eventually stabilizes at moderate-to-large values of $\alpha$, indicating the importance of maintaining sufficient influence from the reference ligand during adaptation.

\section{Conclusion}
Virtual screening plays a crucial role in modern drug discovery. In this work, we investigate how to specialize a pretrained virtual screening model to individual target pockets without full-model retraining and propose PETA. Given a target pocket, PETA retrieves a reference ligand and employs pocket-conditioned diffusion with validity-guided negative mining to construct pocket-aware ligands. Embedding-space mixup and a target-specific ranking objective further transform these ligands into structured supervision for adaptation. By updating only the LayerNorm parameters of the ligand encoder while freezing the remaining model parameters, PETA enables efficient pocket-specific test-time adaptation. Experimental results across diverse benchmarks validate the effectiveness of the proposed approach.
\bibliography{aaai2027}

\end{document}


\maketitle
\appendix

\subsection{Implementation Details}
We set $\alpha=0.9$ for all experiments. For each pocket, we use DiffSBDD with its default pocket-conditioned generation configuration to generate 50 candidate ligands for test-time adaptation.
We use the learned logit scale from the pretrained model, corresponding to an effective temperature of $\tau\approx0.0715$. The model is optimized for 30 adaptation steps per pocket using Adam with $\beta_1=0.9$, $\beta_2=0.999$, $\epsilon=10^{-8}$, and zero weight decay. The learning rate is set to $3\times10^{-3}$ for DUD-E and LIT-PCBA. 
All experiments are conducted on a workstation running Ubuntu 22.04 with Linux kernel 6.8.0. The workstation is equipped with four NVIDIA RTX 6000 Ada Generation GPUs, each with 48\,GB of GPU memory, an AMD Ryzen Threadripper PRO 5965WX CPU with 24 physical cores and 48 threads, and 128\,GB of system RAM. Our implementation uses Python 3.9.25, PyTorch 2.8.0, CUDA 12.8. The primary scientific computing and cheminformatics packages include NumPy 2.0.2, SciPy 1.13.1, scikit-learn 1.6.1, and RDKit 2022.09.3. 
Each experiment is repeated three times using random seeds 0, 1, and 2, and the results are averaged.
We evaluate DrugCLIP and BindCLIP using their official implementations. Since DrugHash does not provide publicly available code, we reimplement it following the original paper. For Drug-TTA, we were unable to reproduce the reported results using its official implementation. Similar concerns have also been raised in its GitHub issue tracker.

\subsection{Details of Evaluation Metrics}
Let $N$ be the number of ligands in a screening library, $A$ the number of active ligands, and $s(m)$ the predicted score of ligand $m$.

\paragraph{AUROC.}
AUROC~\cite{DBLP:journals/prl/Fawcett06} evaluates the global ordering of active and inactive ligands. It is equivalent to the probability that an active ligand receives a higher score than an inactive ligand:
\begin{equation}
\begin{aligned}
    \mathrm{AUROC}
=&
\Pr\!\left(s(m^{+})>s(m^{-})\right)\\&
+\frac{1}{2}\Pr\!\left(s(m^{+})=s(m^{-})\right).
\end{aligned}
\end{equation}
Therefore, $0.5$ represents random discrimination and $1.0$ represents a perfect global ranking.

\paragraph{BEDROC.}
AUROC treats all ranking positions equally, whereas practical virtual screening primarily examines compounds near the top of the list. BEDROC~\cite{DBLP:journals/jcisd/TruchonB07} addresses this difference by exponentially discounting active ligands appearing at lower ranks. Following the DrugCLIP protocol, we compute $\mathrm{BEDROC}_{80.5}$. For active-ligand ranks $\{a_i\}_{i=1}^{A}$, define
\begin{equation}
\begin{aligned}
    &S
=
\frac{1}{A}
\sum_{i=1}^{A}
\exp\!\left(-80.5\,\frac{a_i}{N}\right),
\\
&S_{\mathrm{rand}}
=
\frac{1}{N}
\sum_{j=1}^{N}
\exp\!\left(-80.5\,\frac{j}{N}\right).
\end{aligned}
\end{equation}
The relative early enrichment is $\mathrm{RIE}=S/S_{\mathrm{rand}}$, which is normalized as
\begin{equation}
\mathrm{BEDROC}_{80.5}
=
\frac{\mathrm{RIE}-\mathrm{RIE}_{\mathrm{worst}}}
{\mathrm{RIE}_{\mathrm{best}}-\mathrm{RIE}_{\mathrm{worst}}}.
\end{equation}
Here, $\mathrm{RIE}_{\mathrm{best}}$ and $\mathrm{RIE}_{\mathrm{worst}}$ correspond to placing all active ligands at the beginning and end of the ranked list, respectively. A larger BEDROC value indicates better recovery of active ligands at early ranks.

\paragraph{Enrichment Factor.}
EF~\cite{DBLP:journals/jcisd/TruchonB07} directly measures the improvement over random selection within a specified top-ranked fraction. At a cutoff of $x\%$, let $N_{x\%}$ be the number of selected ligands and $A_{x\%}$ the number of actives among them. The enrichment factor is
\begin{equation}
\mathrm{EF}^{x\%}
=
\frac{A_{x\%}/N_{x\%}}{A/N}.
\end{equation}
An EF of $1$ corresponds to random selection, while a larger value indicates stronger enrichment.

\paragraph{Pairwise Ranking Accuracy.}
Let $\mathcal{E}$ denote the set of ligand-pair edges and $y_{ij}\in\{-1,1\}$ the experimental ordering between ligands $i$ and $j$ derived from the corresponding $\Delta\Delta G$ relation. Given their predicted scores $\hat{s}_i$ and $\hat{s}_j$, pairwise ranking accuracy~\cite{passaro2025boltz} is defined as
\begin{equation}
\mathrm{Acc}_{\mathrm{pair}}
=
\frac{1}{|\mathcal{E}|}
\sum_{(i,j)\in\mathcal{E}}
\mathbb{I}\!\left[
\operatorname{sign}(\hat{s}_i-\hat{s}_j)=y_{ij}
\right].
\end{equation}
It measures the proportion of ligand pairs whose predicted ordering agrees with the experimental relation.

\paragraph{Kendall Rank Correlation.}
We denote the Kendall rank correlation coefficient~\cite{kendall1938new} by $\mathcal{K}$, which measures the agreement between the model-induced global ranking and the experimental free-energy ordering:
\begin{equation}
\mathcal{K}
=
\frac{N_{\mathrm{con}}-N_{\mathrm{dis}}}
{N_{\mathrm{con}}+N_{\mathrm{dis}}},
\end{equation}
where $N_{\mathrm{con}}$ and $N_{\mathrm{dis}}$ denote the numbers of concordant and discordant ligand pairs, respectively. The coefficient ranges from $-1$ to $1$, with a larger value indicating better global ranking consistency.

\subsection{Runtime Analysis}

To evaluate the computational efficiency of PETA, we report the average ligand-generation and model-adaptation times per target pocket. For each pocket, 50 candidate ligands are generated for adaptation. As reported in Table~\ref{tab:runtime}, the complete process takes 52.99 seconds on DUD-E and 57.45 seconds on LIT-PCBA. Model adaptation itself requires less than two seconds on both datasets.

\begin{table}[h]
\centering
\begin{tabular}{lccc}
\toprule
Dataset & Generation (s) & Adaptation (s) & Total (s) \\
\midrule
DUD-E & 51.09 & 1.90 & 52.99 \\
LIT-PCBA  & 55.72    & 1.73  & 57.45    \\
\bottomrule
\end{tabular}
\caption{Average runtime per target pocket across different datasets.}\label{tab:runtime}
\end{table}

\subsection{Similarity Analysis on LIT-PCBA}

\begin{figure}[ht]
    \centering
    \includegraphics[width=0.8\linewidth]{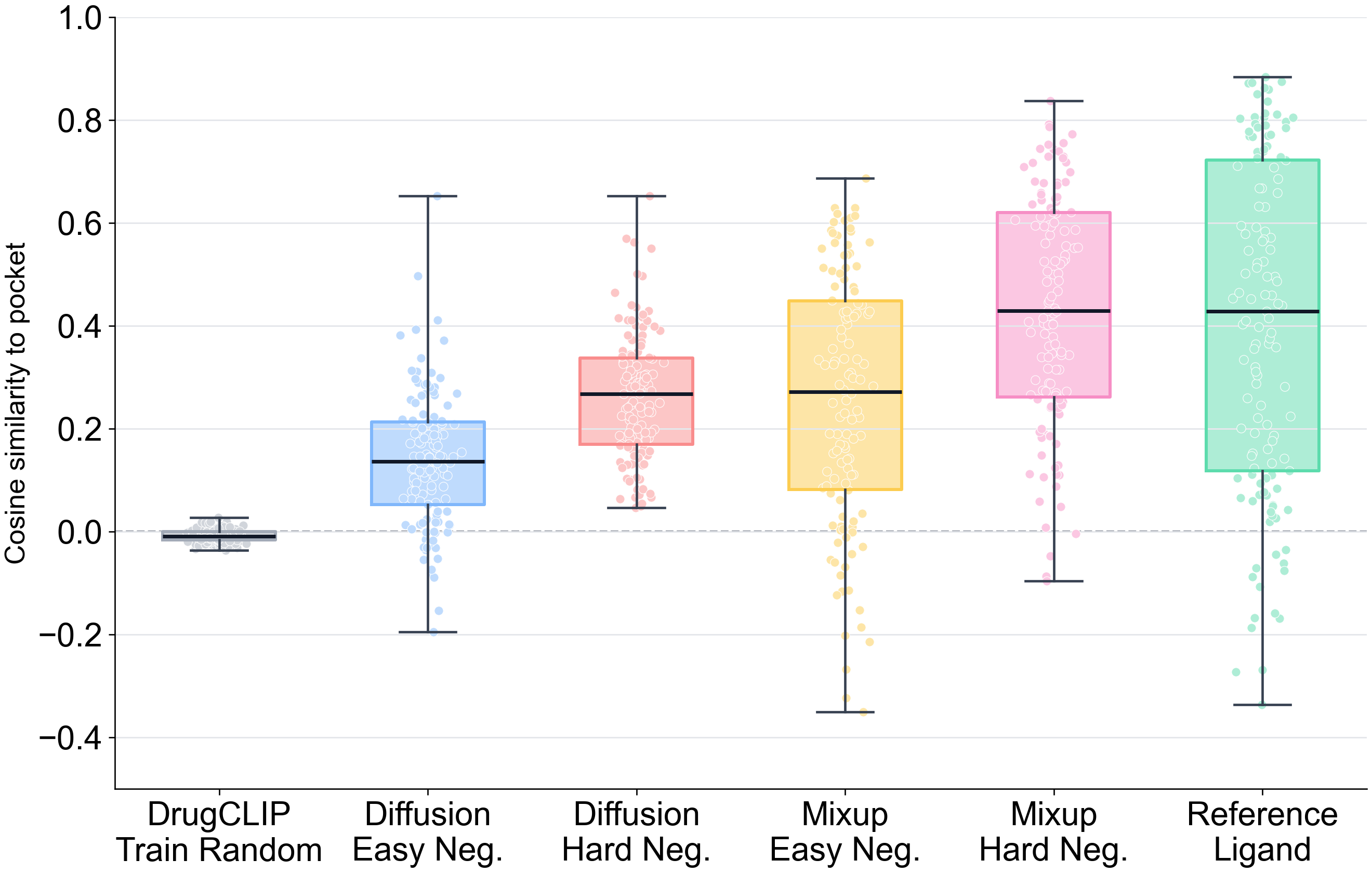}
    \caption{Distributions of per-pocket average pocket--ligand cosine similarities on LIT-PCBA.}
    \label{fig:box-pcba}
\end{figure}

The main text reports per-pocket average pocket--ligand cosine similarities on DUD-E. Here, we provide the corresponding results on LIT-PCBA in Figure~\ref{fig:box-pcba}.

\subsection{Effect of the Mixup Ratio}

\begin{figure}[htpb!]
    \centering
    \begin{subfigure}[t]{0.48\linewidth}
        \centering
        \includegraphics[width=\linewidth]{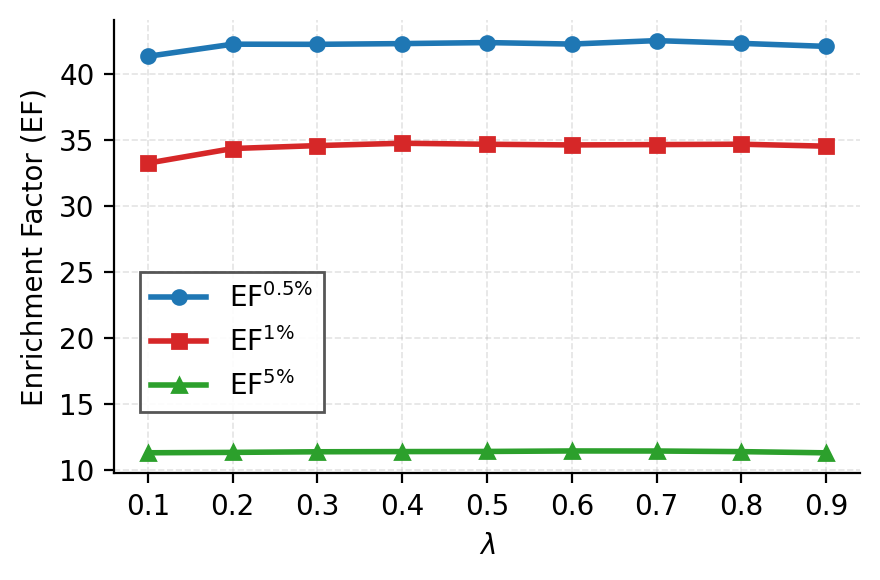}
        \caption{DUD-E}
    \end{subfigure}
    \begin{subfigure}[t]{0.48\linewidth}
        \centering
        \includegraphics[width=\linewidth]{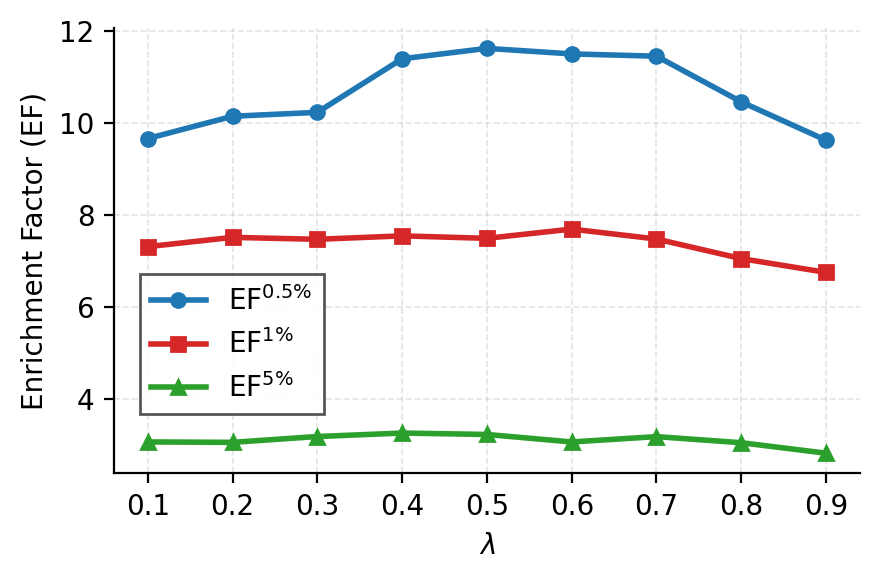}
        \caption{LIT-PCBA}
    \end{subfigure}
    \caption{Effect of the Mixup ratio $\lambda$ on EF scores.}
    \label{fig:mix}
\end{figure}

In the main experiments, we set the Mixup ratio to $\lambda=0.5$. Here, we vary $\lambda$ from 0.1 to 0.9 in increments of 0.1. The resulting EF scores on DUD-E and LIT-PCBA are reported in Figure~\ref{fig:mix}. Performance on DUD-E remains stable across all tested ratios, while LIT-PCBA shows consistently strong performance between 0.4 and 0.7. This suggests that balanced interpolation provides informative intermediate ranking signals and supports the effectiveness of embedding-space Mixup.

\bibliography{aaai2027}